# A neuromorphic hardware framework based on population coding

Chetan Singh Thakur, Tara Julia Hamilton, Runchun Wang, Jonathan Tapson and André van Schaik
The MARCS Institute, University of Western Sydney, Kingswood 2751, NSW, Australia
Email: C.SinghThakur@uws.edu.au

*Abstract*—In the biological nervous system, large neuronal populations work collaboratively to encode sensory stimuli. These neuronal populations are characterised by a diverse distribution of tuning curves, ensuring that the entire range of input stimuli is encoded. Based on these principles, we have designed a neuromorphic system called a Trainable Analogue Block (TAB), which encodes given input stimuli using a large population of neurons with a heterogeneous tuning curve profile. Heterogeneity of tuning curves is achieved using random device mismatches in VLSI (Very Large Scale Integration) process and by adding a systematic offset to each hidden neuron. Here, we present measurement results of a single test cell fabricated in a 65nm technology to verify the TAB framework. We have mimicked a large population of neurons by re-using measurement results from the test cell by varying offset. We thus demonstrate the learning capability of the system for various regression tasks. The TAB system may pave the way to improve the design of analogue circuits for commercial applications, by rendering circuits insensitive to random mismatch that arises due to the manufacturing process.

## I. INTRODUCTION

Semiconductor technology has evolved from discrete single transistors of the 1960's to multi-billion-transistor microprocessors and memory chips of today. This exponential growth in circuit density follows the famous Moore's law, which states that the density of transistors doubles every two years [1]. In the last two decades, IC technology has advanced to nanometre fabrication process technologies. Many physical and quantum mechanical effects that were not relevant in larger process technologies become significant in nanometre designs [2]. These effects lead to problems such as high levels of electrical noise, process mismatch, interconnect bottlenecks, high element failure rate, and power limitations. The result is a serious risk of suboptimal designs and thus poor performance and poor manufacturing yield. These issues render traditional approaches to analogue IC design inadequate and create significant challenges in the field of design technology. Additionally, analogue circuits are more prone to failure in nanometre designs when compared to digital circuits because of their dependence on slight variations of the process, severe impact of noise and leakage currents, influence of external unknown fields and susceptibility to slight changes in layout. These effects may be minimised by increasing device size, however, this increases the size of an IC and hence can be prohibitive in large system-on-chip (SOC) designs [3][4].

Neurobiological processing systems, such as the brain, are remarkable computational devices. Despite their slow speed, they outperform today's modern computers in various tasks such as vision, audition, and motor control. Issues such as cell death and non-homogeneity of neurons in a neurobiological system can be considered equivalent to transistor failure and device mismatch in an IC, respectively. Over the course of evolution, biological systems have evolved to cope with these issues to ensure survival. A set of neurons works collectively and distributively to encode information in the nervous system. Each neuron in such a population has a distribution of responses over some set of input stimuli. The individual neuronal firing rates vary nonlinearly according to the input, and allow decoding of the input value by linearly combining the response of many neurons [5]. In a similar manner, the neuromorphic system that we have described encodes the input stimulus using a large pool of nonlinear neurons, and decodes the desired function by linearly combining responses of neurons. Neuromorphic systems offer an attractive alternative to conventional technology, and have enormous potential for future artificial information processing and behaviour systems [6].

Various applications such as sensor networks, military applications and aerospace require electronic systems with small area, high speed, small weight, and low power consumption. Thus, it is imperative to implement customised neural networks in hardware rather than in software [7]. Analogue neural network hardware is preferable to its digital counterpart in systems requiring small area, low power consumption. Moreover, the former is advantageous owing to high speed resulting from asynchronous updating, and ease of interface with a large set of real world sensors, which are themselves analogue. However, a major drawback of analogue implementation is random device mismatch. We have addressed this issue in our work and exploited random device mismatch as a constructive feature, instead of avoiding it as a bane.

In this paper, we present a novel IC architecture called a Trainable Analogue Block (TAB) that incorporates neuromorphic principles such as low power consumption, fault tolerance and adaptive learning. To our knowledge, our work is the first of its kind to present measurement results of the TAB, which employs random device mismatch to implement a neural network. The TAB implementation is based on the LSHDI (Linear Solutions of Higher Dimensional Interlayers) framework [8], which will be explained in detail in section III. We present the measurements of the building blocks of our TAB hardware implementation in section IV, algorithm for offline learning in section V, mathematical proof for importance of heterogeneity of tuning curves in section VI, and capability of the TAB to learn various regression tasks in section VII. Section VIII presents a comparison of our work to previously published works and the conclusions of our study.

In contrast to existing analogue circuits, TAB architecture embraces random device mismatch. Thus, the reduced device matching in newer technologies serves as an advantage, rather than something that needs to be engineered out of the design. A further significant advantage of this approach is that once manufactured, the same TAB can be reused for many different purposes. The same architecture can be used in different manufacturing technologies, as it can be trained after fabrication to perform a desired operation. This will lead to a significantly reduced design cycle for analogue circuits, with an associated reduction in design cost, and a speed-up of the technological progress. The TAB framework may pave the way for a new kind of circuit paradigm, called stochastic electronics, which will use hardware variability to achieve their engineering goal [9].

## II. NEURAL POPULATION CODING

Biological neurons encode input stimuli such as motion, position, colours, and sound into neuronal firing. The encoded information is represented by a set of neurons in a collective and distributed manner, referred as population coding. In population coding, the firing rate or the rate code of individual neurons governs information encoding. Population coding is robust to neuronal cell damage, as the information is encoded across a large set of neurons [10]. As examples of rate coding, neurons in monkeys, cricket, barn owl, cats, bats and rats encode direction of arm movements [11], direction of a wind stimulus [12], direction of a sound stimulus [13], saccade direction [14], echo delay [15] and position of the rat in its environment [16] respectively. The tuning curve of a neuron is defined as its average firing rate as a function of input stimulus intensity. Various neuronal tuning curves have been identified, such as the cosine tuning curve of motor cortical neurons [17], Gaussian tuning curves of cortical V1 neurons, and sigmoidal tuning curve of stereo V1 neurons. In a similar manner, we have encoded physical quantities into population of neurons by their tuning curves instead of individual spikes in the TAB framework. In our system, inputs are voltage signals, which could be outputs from an array of sensors representing physical quantities of the world. The input stimulus is encoded via the tuning curves of an ensemble of neurons, a phenomenon referred as population encoding.

Neurons within the same cortical column have highly heterogeneous responses to the same input stimulus. The heterogeneity of neuronal responses has been thought to be beneficial for sensory coding when stimuli are decoded from the population response [18][19]. The shape of tuning curves of individual neurons, and the heterogeneity of neuronal responses affect the quality of population coding and the accuracy of information processing in the cortex [20]. We have adapted a similar concept of using heterogeneous population of neurons in a TAB. The significance of heterogeneity of tuning curves is presented in detail in section VI.

As an ensemble of neurons encodes information, decoding the full population response requires procedures for combining the firing rates of many neurons into a population ensemble estimate. One of the popular reconstruction methods known as population vector method was developed by Georgopoulos and collaborators, for coding of the direction of arm movement in monkeys [11]. Abbott et al have discussed various decoding methods, some of which are complex methods based on statistical approaches and use response probabilities, such as Maximum Likelihood Estimation, Bayesian Estimation. Other methods use response tuning curves, such as Least Square Estimation, Projection Method, Vector Method, and Optimal Linear Estimation [21]. In general, each neuron contributes a basis function in this space of variables whenever it fires, and the best estimate of the physical variables is computed from the sum of these functions weighted by the spike rate occurring in each neuron. In our TAB system, we have used a similar approach to decode the stimulus, which is explained in section V.

## III. TAB FRAMEWORK

The TAB framework (Fig. 1) utilises the LSHDI (Linear Solutions of Higher Dimensional Interlayers) principle similar to neural population coding. One of the earliest work in the neural network community, which is based on the LSHDI principle was the Functional-link net computing (FLNN) proposed by Pao et al in 1992 [22]. Similar work was proposed by Schmidt et al in the same year, which however did not attract much attention [23]. In 2006, Huang et al proposed a similar concept as Extreme Learning Machine (ELM) [24], which has attracted widespread attention in the neural network community. The Neural Engineering Framework (NEF) [25] is another example of a network based on the LSHDI principle, which performs spike-based computation and is quite popular in the neuromorphic engineering community.

LSHDI networks are represented as having three layers of neurons – input, hidden and output layers, in a feed-forward structure [8]. These networks differ from similar neural network architectures in several ways – (i) the hidden layer is usually much larger than the input layer, (ii) the connections between the input layer and the hidden layer are randomly generated, and (iii) the connections do not change during the network training. As a result, the inputs are randomly projected from their original input dimensionality to a nonlinear hidden layer of neurons of a much higher dimensionality. Input data points, which are not linearly separable in their current space, allow a linear hyperplane in the higher dimensional space that approximates a desired function as a regression solution, or represents a classification boundary for the input-output relationship. The output layer neurons need therefore compute only a linearly weighted sum of the hidden layer values in order to solve the problem, hence the name Linear Solutions of Higher Dimensional Interlayers (LSHDI) [8]. These linear weights are determined analytically by calculating the product of the pseudoinverse of the hidden layer activations with the desired output values [26].

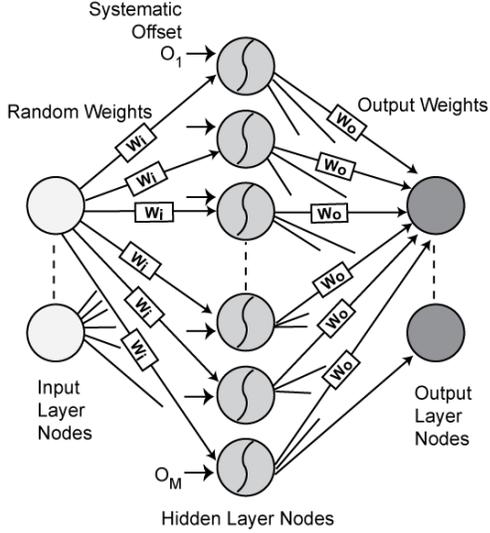

**Fig. 1. Architecture of the TAB framework.** The input layer neurons/nodes are connected to a larger number of non-linear hidden layer neurons via random weights and controllable offsets, $O_1$ to $O_M$. The connections from the hidden layer neurons to the output neurons are linear, with trainable weights, and the output neurons generate a linear sum of their inputs.

## IV. BUILDING BLOCKS OF THE TAB IC

We have designed a test cell, comprising of a Hidden Neuron block and an Output Weight block of the proposed TAB architecture in 65nm technology. Additionally, the TAB architecture allows us to design a major part of the circuit with the lowest possible feature size in order to maximise mismatch among the transistor parameters, because device mismatch is inversely proportional to device area. More importantly, we have added an extra controllable offset (Fig. 1) for each hidden layer neuron to ensure that each hidden neuron performs a different nonlinear operation to the input. This systematic offset is fixed but different for each neuron. Systematic offset may not be required if there is sufficient random variation among transistors to produce a distinct tuning curve for each neuron, but is a failsafe method to spread the tuning curves of the neurons.

The first version of a TAB that we present here has a simple configuration, with a single input voltage and a single output current (single input-single output, SISO). In this section, the VLSI implementation of the major building blocks of the TAB, namely the Hidden Neuron and the Output Weight are described.

### A. Hidden Neuron

Neuroscientists have clearly demonstrated that individual biological neurons respond selectively to various stimuli like sound, motion, images and so on [27]. Each neuron has a distinct tuning curve, which is found by presenting varied input stimuli to the neuron and recording its firing rate. Each neuron encodes the input stimuli according to its tuning curve. Similarly, each artificial neuron on our chip encodes input differently according to its distinct tuning curve.

In the TAB system described here, a simple neuronal tuning curve is implemented using a differential pair, which performs a hyperbolic tangent (*tanh*) nonlinear operation on its input, similar to sigmoidal tuning curve of stereo V1 neurons in the cortex. The circuit is illustrated in Fig. 2A. $M_1$ and $M_2$ constitute the differential pair, and the sharing of currents between $M_1$ and $M_2$ depends on their respective gate voltage, $V_{in}$ (input voltage) and $V_{ref}$ (constant reference voltage). If all MOSFETs ('metal–oxide–semiconductor field-effect transistors') are operating in weak-inversion and at saturation, with the slope factor, $n$ ranging from 1.1 to 1.5, then currents in $M_1$ and $M_2$ transistors can be approximated as:

$I_1 = I_b[exp(V_{in}/nU_T)] / [exp(V_{in}/nU_T) + exp(V_{ref}/nU_T)]$ (1)

$I_2 = I_b[exp(V_{ref}/nU_T)] / [exp(V_{in}/nU_T) + exp(V_{ref}/nU_T)]$ (2)

With ideal transistors, the output currents, $I_1$ and $I_2$, are a function of the input differential voltage between $V_{in}$ (ramp input) and $V_{ref}$ (constant input) and their difference is identical to the mathematical *tanh* function. The current $I_1$ saturates to the maximum bias current if $V_{in}$ is higher than $V_{ref}$ by more than 4 $U_T$ (100 mV), where $U_T$ is the thermal voltage. In the TAB system, each neuron receives a systematically different $V_{ref}$, which results in a different nonlinear curve for each neuron. The fact that the transistors are not ideal, as assumed in (1) and (2) results in further deviations from the *tanh* curve, as shown in Fig. 2B. $I_{tanh}$ is copied from $I_1$ using a current mirror that connects to the Output Weight block. $V_b$ is the voltage at the $M_3$ transistor that sets the bias current, typically in the range of a few nanoamperes (nA).

Fig. 2B shows the tuning curves of a single hidden neuron while varying $V_{ref}$. In the actual TAB, each neuron will have only one such tuning curve depending on its own $V_{ref}$, and dependent on process variations such as offset mismatch between the transistors in the differential pairs, bias current mismatch due to variability in M$_3$ and current mirror mismatch. This illustrates the variation that can be achieved at each hidden neuron by changes in $V_{ref}$ alone.

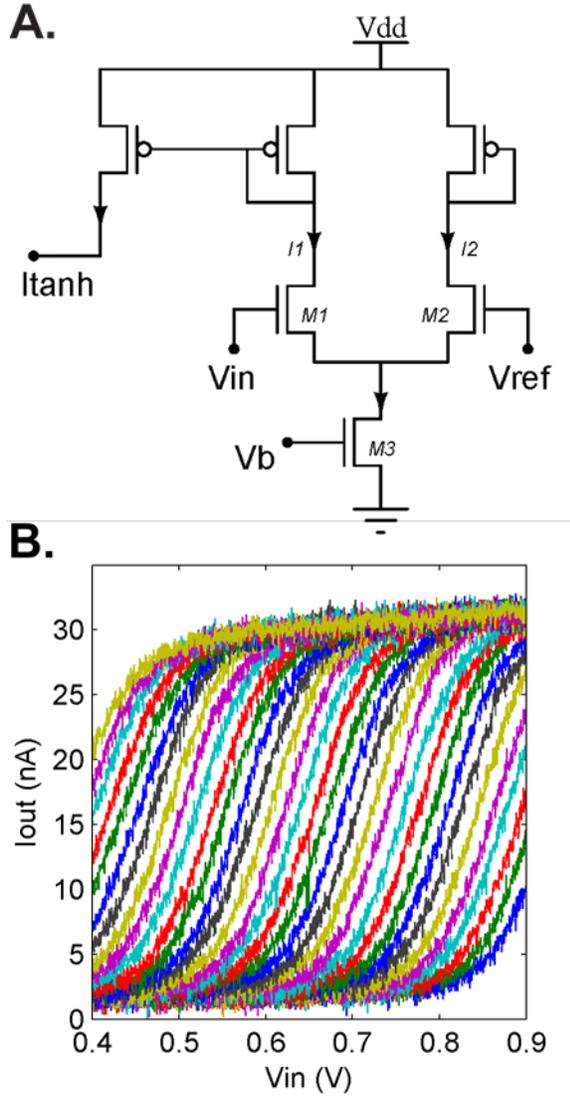

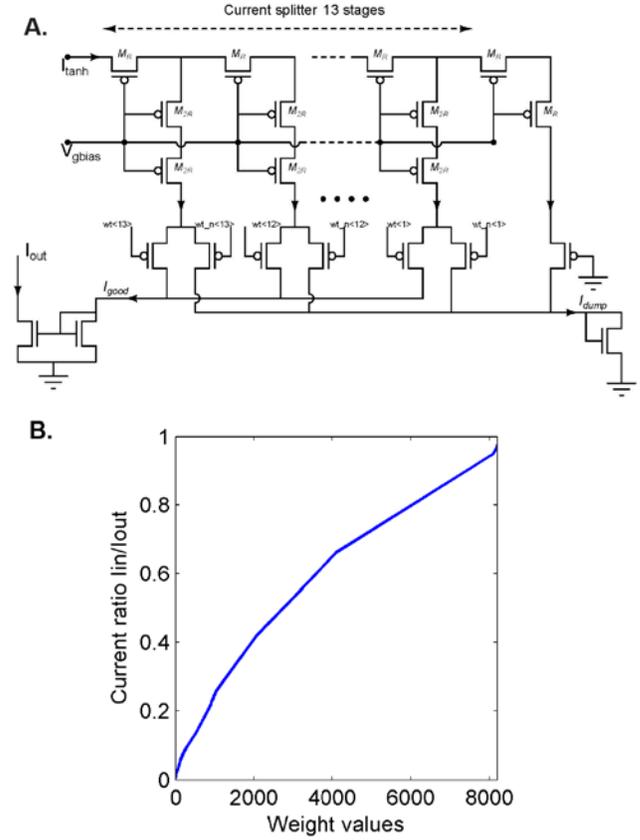

Fig. 2. **Hidden Neuron Block.** This block implements the *tanh* nonlinear activation function for the TAB framework. **A.** Schematic of the Hidden Neuron block. **B.** Measured tuning curves of the hidden neuron as a function of $V_{in}$. Each curve corresponds to different offset $V_{ref}$.

*B. Output Weight*

In the LSHDI framework, there exists a linear relationship between the hidden layer and the output layer. These layers are connected via the Output Weight block. In our TAB SISO chip, the weight is controlled by a binary number. Using simulations, we have found that 11-bits per weight are sufficient for learning various functions. Using more than 11-bits per weight does not improve learning significantly. In the designed test cell, however, we have used a 13-bit binary number for testing purposes. This binary number controls output weights by controlling the amount of current that flows from the hidden layer neurons to the output layer neurons. We have implemented binary weighted connections using a splitter circuit (Fig. 3A) [28].

Fig. 3. **Output Weight Block. A.** Schematic of the Output Weight block. Splitter circuit consisting of M$_R$ and the two M$_{2R}$ transistors form the R2R network, which gets repeated 13 times in the block. The octave splitter is terminated with a single M$_R$ transistor. **B.** Measured current profile of the Output Weight block as a function of the binary weights (13 bits) of the Hidden Neuron block.

Each current branch is controlled through a digital binary switch. The input current, $I_{tanh}$, which is the output of the neuron block, is divided successively to form a geometrically-spaced series of smaller currents. At each branch, a fixed fraction of the current is split off, while the remainder continues to later branches. The last stage is sized to terminate the line as though it were infinitely long. The current splitter principle accurately splits currents over 20 octaves, spanning from weak to strong inversion of transistors, dependent only on the effective device geometry. M$_R$ and the two M$_{2R}$ transistors form an R2R network, and the octave splitter is terminated with a single M$_R$ transistor. The splitter has N stages; the current at the k$^{th}$ stage is ($I_{tanh}/2^k$). The final current is the same as the penultimate current. The transistor sizes of M$_R$ and M$_{2R}$ are equal. The reference voltage for the p-FET gates in the splitter is the master bias voltage $V_{gbias}$ [28]. The lower half of the R2R block has two transistor switches that act as a binary synapse for every branch, and route the branch current to either useful current, $I_{good}$, or to current that goes to ground, $I_{dump}$. $I_{good}$ is mirrored to generate presynaptic current $I_{out}$ for the output layer neuron. Fig. 3B shows the measured output current of the Output Weight block with respect to various binary weights.

## V. OFFLINE LEARNING OF THE TAB IC

Learning in the TAB framework is achieved by computing output weights to train the system for desired regression/classification tasks. Briefly, the LSHDI framework determines the output weights (between the large hidden layer and linear output neurons) analytically by calculating the product of the pseudoinverse of the hidden layer activations with the target outputs [29].

A novel algorithm used for offline learning on the TAB IC is discussed here. Let us consider a three-layer feed-forward TAB network with $L$ number of hidden neurons. Let $G(.,.,.)$ be a real-valued function so that $G(w_i^{(1)}, b_i^{(1)}, o_i^{(1)}, x)$ is the output of the $i^{th}$ hidden neuron, corresponding to the input vector $x \in \mathbb{R}^m$ and the random input weight vector $w_i^{(1)} = (w_{i1}^{(1)},... w_{im}^{(1)})$, where $w_{is}^{(1)}$ is the weight of the connection between the $i^{th}$ hidden neuron and $s^{th}$ neuron of the input layer. Random bias vector $b_i^{(1)} \in \mathbb{R}$ and the random input weight vector $w_i^{(1)}$ both arise due to random mismatch of the transistors. Systematic offset $o_i^{(1)} \in \mathbb{R}$ is added to make sure each neuron exhibits distinct tuning curve, which is an essential requirement for learning in LSHDI framework, discussed in detail in section VI. The output function $f(.)$ is given by:

$$f(x) = \sum_{i=1}^{L} w_i^{(2)} G(w_i^{(1)}, b_i^{(1)}, o_i^{(1)}, x) \quad (3)$$

where, $w_i^{(2)} = (w_{1i}^{(2)},... w_{ki}^{(2)}) \in \mathbb{R}^k$ is the weight vector where $w_{ji}^{(2)} \in \mathbb{R}$ is the weight connecting the $i^{th}$ hidden neuron with the $j^{th}$ neuron of the output layer. Here, $G(.,.,.)$ takes the following form:

$$G(w_i, b_i, x) = g(w_i^{(1)}.x + b_i^{(1)} + o_i^{(1)}) \quad (4)$$

where, $g: \mathbb{R} \rightarrow \mathbb{R}$ is the activation function.

Suppose, for a training data set $\{(x_n, y_n)\}_{n=1,2..C}$, $x_n = (x_{n1},..., x_{nm}) \in \mathbb{R}^m$ denotes the input vector, $y_n = (y_{n1},..., y_{nk}) \in \mathbb{R}^k$ is its corresponding output vector, and $C$ is the total number of input data patterns. Let the values of the input weight vectors, $w_i^{(1)} \in \mathbb{R}^m$, the bias, $b_i^{(1)} \in \mathbb{R}$, be randomly assigned and $o_i^{(1)} \in \mathbb{R}$, be assigned systematically. Then, the standard TAB framework with $L$ number of hidden neurons approximates the input samples with zero error if and only if there exists $w_i^{(2)} \in \mathbb{R}^k$ such that:

$$y_n = \sum_{i=1}^{L} w_i^{(2)} G(w_i^{(1)}, b_i^{(1)}, o_i^{(1)}, x_n) \text{ where, } n = 1,2,..C \quad (5)$$

The above set of equations can be rewritten in the following matrix form as:

$$HW^{(2)} = Y \quad (6)$$

where,

$$H_{MxL} = \begin{Bmatrix} G(w_1^{(1)}, b_1, o_1, x_1) & \dots \dots & G(w_L^{(1)}, b_L, o_L, x_1) \\ \vdots & & \vdots \\ \vdots & & \vdots \\ \vdots & & \vdots \\ G(w_1^{(1)}, b_1, o_1, x_C) & \dots \dots & G(w_L^{(1)}, b_L, o_L, x_C) \end{Bmatrix} \quad (7)$$

$$W^{(2)}_{LxK} = \begin{Bmatrix} w_1^{(2)} \\ \vdots \\ \vdots \\ \vdots \\ w_L^{(2)} \end{Bmatrix}, \quad Y_{CxK} = \begin{Bmatrix} y_1 \\ \vdots \\ \vdots \\ y_C \end{Bmatrix} \quad (8)$$

Here, the $i^{th}$ column of $H$ will be the output of the $i^{th}$ hidden neuron for all the input training data samples ($x_1,..., x_m$). Further, the matrix $H$ need not be a square matrix. Under the assumption that the activation function $g(.)$ is infinitely differentiable, it has been shown that for fixed input weight vectors, $w_i^{(1)}$, and biases, $b_i^{(1)}$, $o_i^{(1)}$, the least squares solution $W^{(2)}$ for the matrix (6) is:

$$W^{(2)} = H^+ Y \quad (9)$$

where, $H^+$ is the Moore-Penrose generalised pseudoinverse of the matrix $H$.

The output weight calculation can be summarised as follows:

*Input:* Training set $\{(x_n, y_n)\}_{n=1,2..C}$, $x_n \in \mathbb{R}^m$ and $y_n \in \mathbb{R}^k$, $L$ is the number of hidden neurons, and the activation function is $g(.)$

1. For $i = 1,2,...L$, randomly assign the input weight vector $w_i^{(1)} \in \mathbb{R}^m$, random bias $b_i^{(1)} \in \mathbb{R}$ and systematic offset $o_i^{(1)} \in \mathbb{R}$.
2. Determine the matrix $H$ defined by the (7).
3. Calculate $H^+$.
4. Calculate the output weights matrix as $W^{(2)} = H^+ Y$, where $Y$ is given by (8).

*Output:* Network with the determined output weight vectors $w_i^{(2)} \in \mathbb{R}^k$ for the randomly chosen weight vectors $w_i^{(1)} \in \mathbb{R}^m$, bias $b_i^{(1)} \in \mathbb{R}$ and systematic offset $o_i^{(1)} \in \mathbb{R}$ for $i = 1,2,...L$ will compute the estimated output value $\hat{y}$ for any input test sample $x \in \mathbb{R}^m$ using the following formula:

$$\hat{y} = \sum_{i=1}^{L} w_i^{(2)} g(w_i^{(1)}.x + b_i^{(1)} + o_i^{(1)}) \quad (10)$$

## VI. IMPORTANCE OF HETEROGENEITY OF TUNING CURVES

It has been demonstrated that in a neurobiological system, individual neurons exhibit highly heterogeneous responses when presented with the same stimuli. This heterogeneity has been shown to improve the information encoded in the neuronal population activity by decreasing the neuronal correlations. Diversity of orientation tuning curve profiles of individual neurons proves beneficial for sensory coding when stimulus orientation is decoded from the population response [18]. Similarly, we show that the tuning curves of neurons in our TAB framework should be heterogeneous so as to have the highest information encoding capacity.

Let us revisit equation (6) and find $W^{(2)}$ analytically:

$$Y = HW^{(2)}$$

$$H^T Y = H^T H W^{(2)}$$

$$(H^T H)^{-1} H^T Y = W^{(2)}$$

Then estimated output,

$$\hat{Y} = HW^{(2)} = H((H^T H)^{-1} H^T Y)$$

The error of the system, which is the difference between estimated $\hat{Y}$ and actual $Y$, is given as:

$$E = (Y-\hat{Y})$$

$$E = (Y-H(H^T H)^{-1} H^T Y)$$

$$E = Y(I-H(H^T H)^{-1} H^T)$$

where, $I$ is the identity matrix. For $E=0$,

$$I = H(H^T H)^{-1} H^T \quad (11)$$

If matrix $H$ is a full column rank matrix, or equivalently, columns of the matrix $H$ are linearly independent, then,

$$I = HH^+ \quad (12)$$

where, $H$ is the measured matrix containing the hidden neuron output across the range of input values and $H^+ = (H^T H)^{-1} H^T$ is the pseudoinverse of a matrix [30]. Equation (11) implies that any input vector $x_i$ is perfectly encoded when the $i^{th}$ row of the RHS matrix is equal to the $i^{th}$ row of the identity matrix $I$. Thus, the encoding capacity of the network is proportional to the number of rows that are equal between the matrices on both sides of equation (11). Mathematically, the more the number of tunning curves of the hidden neurons are independent, the higher is the encoding capacity of the TAB system. Tuning curves of individual neurons are independent in a heterogeneous neuron population, which thus increases the encoding of information [18],[31]. This is a very important observation, and is essential for optimal learning in an LSHDI network. In the next section, we discuss how we can maximise the encoding capacity of the TAB.

***Role of systematic offset:*** When a population of biological neurons is presented an input stimulus, the neuronal responses vary widely owing to variations in neuronal response properties, such as mean firing rate, receptive field location, and stimulus selectivity. Such heterogeneity of responses results in faithful encoding of information covering the whole range of input stimuli. We have tried to create a heterogeneous population of neurons in our TAB system by exploiting randomness (fixed-pattern transistor mismatch) and variability arising due to the fabrication process. An element of risk and uncertainty is present here, as one cannot be certain there would be sufficient mismatch in a particular technology until after manufacturing. For example, older technologies with large feature sizes generally have a low degree of mismatch, which would limit the learning capability of a TAB. This risk is attenuated by introducing a fixed and distinct systematic offset (Fig. 1) for each hidden layer neuron of the TAB. The systematic offset ensures that all tuning curves are distinct and independent, thus improving the encoding capacity of the system. We have found in simulations that if the neuronal tuning curves are too similar, the system requires an extremely large number of hidden neurons to learn even simple functions.

## VII. LEARNING RESULTS

In this section, we describe the learning capability of the TAB system in software using the measured results of the building blocks. We have used the hidden neuron's tuning curves with respect to different $V_{ref}$ voltages (34 in total). The TAB architecture was trained to implement various functions such as *sin* (Fig. 4A), *cube* (Fig. 4B), and *sinc* (Fig. 4C). The learning capability of an LSHDI network depends on the number of hidden layer neurons [8]. As evident from Fig. 4, 34 neurons are sufficient to learn simple functions such as *sin* and *cube*, but for complex function like *sinc* (Fig. 4C), a higher number of neurons is needed for more accurate performance. We have calculated the output weights externally using offline learning method as discussed in section V. The results suggest that the system can be successfully trained to perform the various regression tasks.

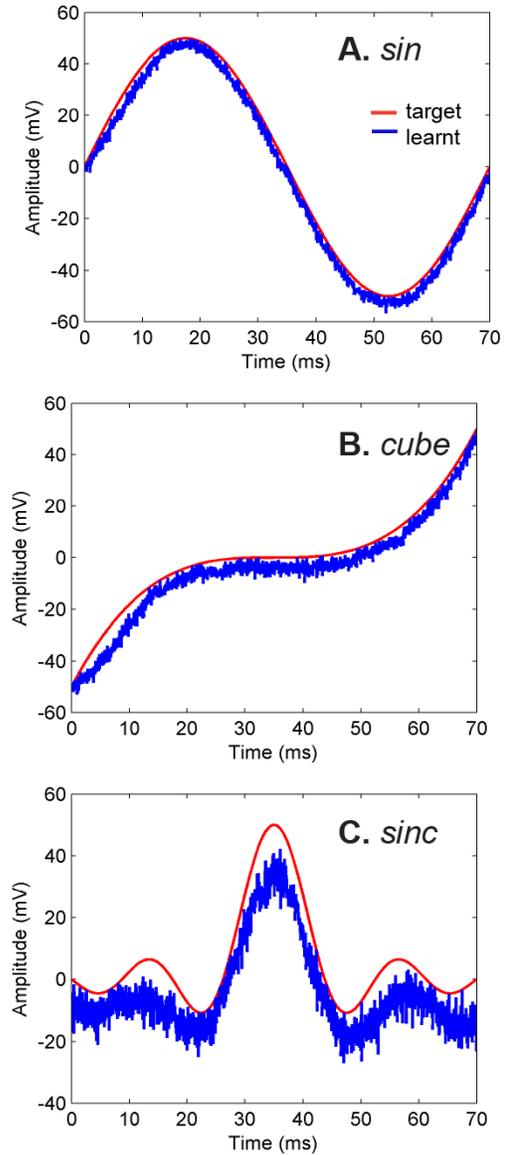

**Fig. 4.** Learning curves for the regression functions – **A.** *sin*, **B.** *cube*, **C.** *sinc*. The *red* curve represents the target function, and the *blue* curve represents learnt function.

## VIII. CONCLUSIONS

In this paper, we have described a novel framework that exploits device mismatch in circuits and performs reliable computation. We have presented measurement results of the building blocks of the TAB IC implementation. Additionally, we have shown the potential learning capability of the TAB system.

In IC technology, random device mismatch is a major problem which always leads to a suboptimal design. It may be minimised to some extent in higher process technology (>100nm) with large device area and good design effort, which however increases production costs significantly. Other research groups have also suggested using random device mismatch in their architecture [32], [33]. However, there are major differences in their architecture compared to ours. Basu *et al* [32] have shown a spiking neuron-based framework which converts analogue input into spikes and translates them into spike rate using counter for each neuron and the rest of the computation is performed in digital controller. Kudithipudi *et al* [33] have developed a memristor-based network in software simulations, which has many practical issues to be considered in chip implementation.

The amount of random mismatch depends on process technology – it is low for a process technology with a large feature size, and vice versa. It also varies from chip to chip. Both [32] and [33] have used software simulations explicitly generates the desired distribution of randomness (variance) which then leads to distinct tuning curves for the entire input range. In their system, the input layer weights of the LSHDI network only rely on this random mismatch obtained from fabrication, which may not be sufficient to generate diverse neuronal tuning curves. In our implementation we avoid this issue by providing an alternative to obtain the systematic offset generated through the resistive polyline, which generates a different $V_{ref}$ voltage for each neuron to ensure that each neuron has a different tuning curve.

An LSHDI network has a large number of hidden neurons. Both previous designs have represented the input variable as a current. Kudithipudi *et al* group have used a resistor before the differential amplifier to convert the total input current to a voltage. They require resistors in the range of few mega-ohms to operate their system in the sub-threshold region. In the sub-threshold region, the differential amplifier will get saturated as the differential voltage goes above $4U_T$ (~100mV). Since an LSHDI network requires a large number of hidden neurons, using many resistors in the design will increase the chip area unrealistically. Also, current would increase as a multiple of the number of inputs and may affect the chip adversely. The above system also uses memristors. This might be a good alternative, but memristors are still in the research phase, and may cause problems in physical realisation [34].

Here, we have presented measurement results of a test cell in a TAB framework. We have shown the learning capability of the TAB system for various regression tasks. Future work will aim to test the learning capability of a complete TAB chip and will include quantification of the random variations across the hidden neurons. The TAB system is designed using neuromorphic principles based on stochastic computation, which has the advantages of low power consumption, adaptability to local change and the ability to learn. This system may help overcome limitations of analogue IC design at low process nodes and will drive the integration process with digital blocks in the same circuit and process node. This may find applications in analogue/digital converters (ADCs) and digital-to-analogue converters (DACs) for submicron mixed signal chips such as those used in mobile processor chips and data acquisition chips. We know that TABs require a large number of hidden layer nodes and connections to and from these nodes; however, unlike custom analogue design, minimum sized transistors can be used and no specialised layout techniques will be required. Furthermore, as the TAB framework desires large random mismatch among devices and as mismatch is inversely proportional to device area, it could lead to significant reductions in chip area and manufacturing costs.